\documentclass{llncs}

\usepackage{times}
\usepackage{url}
\usepackage{latexsym}
\usepackage{mathtools}
\usepackage{hyperref}
\usepackage{algorithm2e}
\usepackage{comment}

\SetKwProg{Fn}{Function}{}{}

\begin{document}

\title{Building Morphological Chains for Agglutinative Languages}

\author{Serkan Ozen\inst{1} \and Burcu Can \inst{2}}
\authorrunning{Burcu Can et al.} 
%
\tocauthor{Serkan Ozen, Burcu Can}
\institute{Department of Computer Engineering \\ 
Middle East Technical University (ODT\"U) \\
Ankara, 06800, Turkey\\
\email{serkan1ozen@gmail.com}
\and
Department of Computer Engineering, Hacettepe University \\ 
Beytepe, 06800, Ankara, Turkey \\
\email{burcucan@cs.hacettepe.edu.tr}
}
\date{}

\maketitle
\begin{abstract}
In this paper, we build morphological chains for agglutinative languages by using a log linear model for the morphological segmentation task. The model is based on the unsupervised morphological segmentation system called MorphoChains~\cite{narasimhan2015unsupervised}. We extend MorphoChains log linear model by expanding the candidate space recursively to cover more split points for agglutinative languages such as Turkish, whereas in the original model candidates are generated by considering only binary segmentation of each word. The results show that we improve the state-of-art Turkish scores by 12\% having a F-measure of 72\% and we improve the English scores by 3\% having a F-measure of 74\%. Eventually, the system outperforms both MorphoChains and other well-known unsupervised morphological segmentation systems. The results indicate that candidate generation plays an important role in such an unsupervised log-linear model that is learned using contrastive estimation with negative samples. 
\end{abstract}

\section{Introduction}
Unsupervised morphological segmentation has been one of the fundamental tasks in natural language processing. Segmentation of words is required normally as a pre-processing task in many natural language processing applications, such as machine translation, question answering, sentiment analysis and so on. One of the main reasons to perform morphological segmentation before applying any natural language processing task is the out-of-vocabulary (OOV) problem. The number of different word forms can be theoretically infinite in agglutinative languages \cite{Hankamer}.

Morphological analysis is also required for some natural language processing tasks. In a full morphological analysis, morphemes are tagged according to their syntactic roles in addition to finding the morpheme boundaries. For example, in order to distinguish the word that is inflected with the negation suffix \textit{ma} (or \textit{me} depending in the vowel harmony) from another word that has a derivational suffix \textit{ma} (or \textit{me}) in Turkish requires a full morphological analysis. Here, we aim to perform morphological segmentation rather than a full morphological analysis. Thus we only aim to find the morpheme segmentation points of each word. 

In this paper, we propose an improvement to the MorphoChains segmentation system~\cite{narasimhan2015unsupervised} by extending the candidate space used in contrastive estimation, thereby covering also agglutinative languages for multiple split points. Normally, log-linear models are supervised. However, using contrastive estimation by shifting the probability mass from the unobserved data (and possibly that are impossible to observe in data) to observed data enables unsupervised learning. Unobserved data is generated with negative sampling using the observed data through some transformations on the observed data (such as transpose, deletion, insertion etc.).

In this paper, rather than extending the probability mass assigned for unobserved data, we target the probability mass assigned for the observed data. For that purpose, we generate more segmentation points (i.e. candidates) for each observed word to extend the observed space. 

Unsupervised models seem to be a good alternative for discovering both orthographic and semantic features of words. We also adopt both orthographic features and semantic features in this paper as proposed in the original model. 

We perform all experiments on publicly available Turkish, English and German datasets provided by Morpho Challenge 2010~\cite{MorphoChallengeWeb}. The evaluation method will be the same with the one used in MorphoChains segmentation system~\cite{narasimhan2015unsupervised}. 

The paper is organized as follows: Section~\ref{related_work} addresses the related work on unsupervised morphological segmentation, section~\ref{model} describes the extended log-linear model, section~\ref{improvements} explains the improvements performed on the original log-linear model, section~\ref{results} presents the experiments and scores for English, Turkish and German along with a discussion over the scores, and finally section~\ref{conclusion} concludes the paper with the potential future work. 

\section{Related Work} 
\label{related_work}

Morphological segmentation is one of the oldest natural language processing tasks that has been excessively studied. 

The oldest works have been usually based on deterministic methods. One of the earliest works is \textit{Linguistica} that is proposed by Goldsmith~\cite{Goldsmith:2001:ULM:972667.972668}. The model is based on Minimum Description Length (MDL) principle, which is deterministic. \textit{Linguistica} employs morphological structures called signatures in order to represent words. Signatures reflect the internal structure of words. Words with similar morphological structure reside in the same signature. For example, ~\textit{\{order, walk\}}-\textit{\{ing, s\}} make a signature that covers words such as \textit{walking, ordering, walks, orders}, and \textit{\{paper, pen\}}-\textit{\{s\}} make another signature that covers \textit{papers, pens}. 

Probabilistic methods have also been used in unsupervised morphological segmentation. Creutz and Lagus~\cite{creutz2002unsupervised} introduce another well-known unsupervised morphological segmentation Morfessor Baseline, the first member of the Morfessor family. One of the versions is based on MDL principle and the other one is based on Maximum Likelihood (ML) estimate. In another member of the same family, Creutz and Lagus~\cite{Creutz05inducingthe}, suggest using priors by converting the model into a Maximum a Posteriori model, thereby introducing another member of the same family, called Morfessor Categories MAP (Maximum A-posterior). Morfessor has been one of the main reference segmentation systems to compare with most of the unsupervised segmentation systems. In this paper, we also compare our extended model with Morfessor Baseline and Morfessor CatMAP. 


Non-parametric Bayesian methods have also been used in segmentation task. Goldwater et al.~\cite{goldwater} present a framework that generates power-laws by using word frequencies. Pitman-Yor Process~\cite{Ishwaran} (the two parameter extension of a Dirichlet Process) is used as a stochastic process in their framework. Snyder and Barzilay~\cite{snyder2008unsupervised} use Dirichlet Process, the simplified version of the Pitman-Yor Process, to induce morpheme boundaries on a bilingual aligned corpus simultaneously by finding the cross-lingual morpheme relations. Lee et al.~\cite{lee2011modeling} address the connection between syntax and morphology in a statistical model. Syntactic knowledge is incorporated in their morphological segmentation system. Their results show that using syntactic information helps in morphological segmentation. 

Some of the systems not only attempt to perform morphological segmentation, but also aim to learn hidden structures behind words. Chan~\cite{chan2006learning} applies Latent Dirichlet Allocation (LDA) to learn morphological paradigms as latent classes. The model assumes that correct segmentations of words are known but morphological paradigms are to be learned. Chan discovers that the final morphological paradigms can be matched with syntactic tags (such as noun, verb etc.). Can and Manandhar~\cite{Can2010} obtain syntactic categories from a context distributional clustering algorithm~\cite{Clark} and learn paradigms by using the the pairs of syntactic categories that have common stems. 

Similar to MorphoChains system, log-linear models have also been utilized in morphological segmentation. Poon et al.~\cite{poon2009unsupervised} suggest using bi-gram morpheme contexts in a log linear model similar to the current study in this paper. In addition to morpheme contexts, Minimum Description Length-inspired (MDL) prior information is also used in their model to keep the lexicon and corpus size small.



In the recent years, deep neural networks are used for learning morphology. Cao and Rei~\cite{cao2016joint} propose a model where word embeddings and segmentation are learned simultaneously. Soricut and Och~\cite{soricut} learn the morphological transformations between words using a high dimensional vector space (i.e. word-embedding space).

In this paper, 200-dimensional neural word embeddings obtained from word2vec~\cite{Mikolov} are also used to capture semantic similarities between words that are derived from each other. 

\section{Model}
\label{model}
\subsection{Model Definition}

In this paper, we extend the MorphoChains\cite{narasimhan2015unsupervised} segmentation system where each word and its morphological roots are represented as a chain structure. For example, \{\textit{walking, walk}\} and \{\textit{undoable, doable, do}\} make morphological chains. In the morphological chain, each word appears in a parent-child relation. Here, \textit{walk} is the parent of \textit{walking}; \textit{doable} is the parent of \textit{undoable}, and \textit{do} is the parent of \textit{doable}. 

In the MorphoChains system, a log-linear model is used to extract the chain structure in an unannotated corpus. The model has a feature vector $\phi$: W $\times$ Z $\rightarrow$ $R^d$ and a corresponding weight vector $\theta$ $\in$ $R^d$, where $W$ denotes words and $Z$ denotes candidates. A candidate is a potential parent set of a word. For example, the word \textit{doors} has the following candidates: (\textit{door, suffix}), (\textit{doo}, suffix), (\textit{do}, suffix), (\textit{rs}, prefix), (\textit{ors},prefix), (\textit{oors}, prefix). Every word and candidate pair has a feature vector associated with it. 

Probability of a word-candidate pair (w,z) is modeled as:
\begin{equation}
P(w,z) = e^{\theta \cdot \phi(w,z)}
\end{equation}
where $w$ is a word and $z$ is a candidate of $w$. Thus, the conditional probability of a candidate given its word is computed by:
\begin{equation}
P(z|w) = \frac{e^{\theta \cdot \phi(w,z)}}{\sum_{z'\in C(w)}{e^{\theta \cdot \phi(w,z)}}}
\end{equation}
where $C(w)$ corresponds to the candidates of $w$. 

The log-linear model proposed in the original paper uses features and their weights in order to learn the underlying segmentation of words. These features are described in the following section.  

\subsection{Features}

Features play a key role in a log-linear model as they represent both orthographic and semantic properties of word-candidate pairs. The features in the model are as follows:

\textbf{Semantic Similarity} is applied by the cosine similarity of a word-parent pair. The cosine similarity is computed by using the word embeddings obtained from word2vec \cite{Mikolov}. The paper indicates that morphologically related word-parent pairs tend to have high cosine similarity. For example, ~\textit{(fly, flying)} pair will have higher cosine similarity when compared to \textit{(flyi, flying)} and this will favor \textit{fly} to be the parent of \textit{flying} rather than having \textit{flyi} as the parent.

\textbf{Affixes} are automatically generated as a list of most frequent affixes in the corpus. In order to build the affix list, each word having a higher frequency than a manually set threshold is analyzed through its potential suffixes and prefixes. All potential suffixes are added into the affix list. If another word in the corpus having the same suffix is met, then the frequency of the suffix is incremented.

\textbf{Affix correlation} shows how related two affixes are in terms of the rate of their common stems. For example ~\textit{(ing, ed)} suffix pair is expected to have a high correlation since many verbs in English can take both of the suffixes. If two affixes share common stems, then they are called neighbor suffixes. Again the same pair ~\textit{(ing, ed)} are called neighbors since they share many common stems. For example, regarding the word \textit{(laughing)} and the suffix \textit{(-ing)}, since another word \textit{(laughed)} exists in the corpus, the parent-candidate pair \textit{(laugh, laughing)} gets a feature stating that \textit{(-ing)} is most probably a suffix which in turn favors \textit{(laugh)} to be a strong candidate for \textit{(laughing)}.

\textbf{Presence in the wordlist} represents whether the parent is seen in the corpus or not. This provides a bias on the likelihood of a parent to be a valid word. This feature assumes that the language is concatenative. 

\textbf{Transformation features} are used for stem changes during affixation. There are three types of transformation features, namely \textit{repeat, delete, modify}. For example, \textit{(running, run)} word-candidate pair has a repeat feature set to 1 due to  the repetition of \textit{n} at the end of the word, and \textit{(deleting, delete)} pair has a delete feature set to 1 due to the deletion of the letter \textit{e} at the end of the word. 

\textbf{Stop features} help to identify whether a parent is the root or not. One of the key features to handle this is the highest cosine similarity between a word and its parents. For example, for the word \textit{flying}, \textit{fly} is more likely to be the base word than \textit{fl} because cosine similarity between \textit{flying} and \textit{fly} is higher than the cosine similarity between \textit{fl} and \textit{flying}.

\section{Improvements to the Model}
\label{improvements}
The model is learned in an unsupervised setting my maximizing the likelihood of observed words in a given corpus. The likelihood of the model for a given unannotated word list $D$ is given as follows in the original paper:
\begin{eqnarray}
\label{likelihood}
L(\theta,D) &=& \prod_{w^*\in D}P(w^*) \nonumber \\
			&=& \prod_{w^*\in D}\sum_{z\in C(w*)}^{} P(w^*,z) \\
          &=& \prod_{w^*\in D}\Big[\frac{\sum_{z\in C(w^*)}^{} e^{\theta\cdot \phi(w^*,z)}}{\sum_{w\in \Sigma*}{\sum_{z\in C(w)}^{} e^{\theta\cdot \phi(w,z)}}}\Big] \\  
\end{eqnarray}
where $\Sigma*$ denotes the alphabet, which is problematic to calculate for all possible unobserved data for a given language. Contrastive estimation is used to apply negative sampling and replace the normalization term with the neighbors of each word. This process creates a large space of unobserved data from which the probability mass will be shifted to observed space and therefore the likelihood will be normalized through the unobserved data. 

\begin{algorithm}[t!]
 \label{algorithm}
 \Fn{RCG (word, candidateList)}{
 \KwData{word}
 \KwResult{candidateList}
   \For{i=word.length-1;$i\geq $word.length-4 and $i \geq $0; i=i-1}{  
     $parent {\leftarrow} $word.substring(0, i)\;
     \If{2 * $parent.length \geq $word.length and $word.length > $2}{
       $candidateList \leftarrow parent$\;
       AddSuffixFeature(parent)\;
       RCG(parent, candidateList)\;
     }
     $parent {\leftarrow} $word.substring(i,word.length)\;
     \If{2 * $parent.length \geq $word.length}{
       $candidateList \leftarrow parent$\;
       AddPrefixFeature(parent)\;
     }
   }
   }
 \caption{RCG (Recursive candidate generation) algorithm}
\end{algorithm}

Here, we have noticed that although candidates play an important role in the model, they are generated by only binary segmentation of each word. For example, the Turkish word \textit{kitap+lar+dan} (from the books) will never have the suffix \textit{lar} in any of its candidates. This holds true for any word with more than one suffix. With this intuition, we aim to increase the candidate space generated from each word by including all possible segmentations of each word, therefore introducing the suffixes in the middle of words as candidates as well.  

In our approach, in order to generate all possible candidates of a word, each binary segmentation of the word is proposed as a candidate. For each candidate stem obtained from the binary segmentation, candidate segmentations are generated again with a binary segmentation. Therefore, a left-recursion is applied for each word in all levels of the binary segmentation in order to generate candidates. In other words, the process is repeated recursively for each candidate stem in each iteration. 

We restrict some candidate generations with some heuristics. Each candidate has a maximum suffix length of $4$. The recursion continues as long as the base word's length is greater than $2$. Another heuristic in the original model is that twice of parent's length must be greater than or equal to the twice of the child word's length. For example, candidates of the word \textit{cars} will be \textit{car, ca, rs, ars}. Words \textit{(c, s)} are detained from being candidates since they do not meet any of the heuristics. 

\begin{figure*}[t!]
  \includegraphics[width=\linewidth]{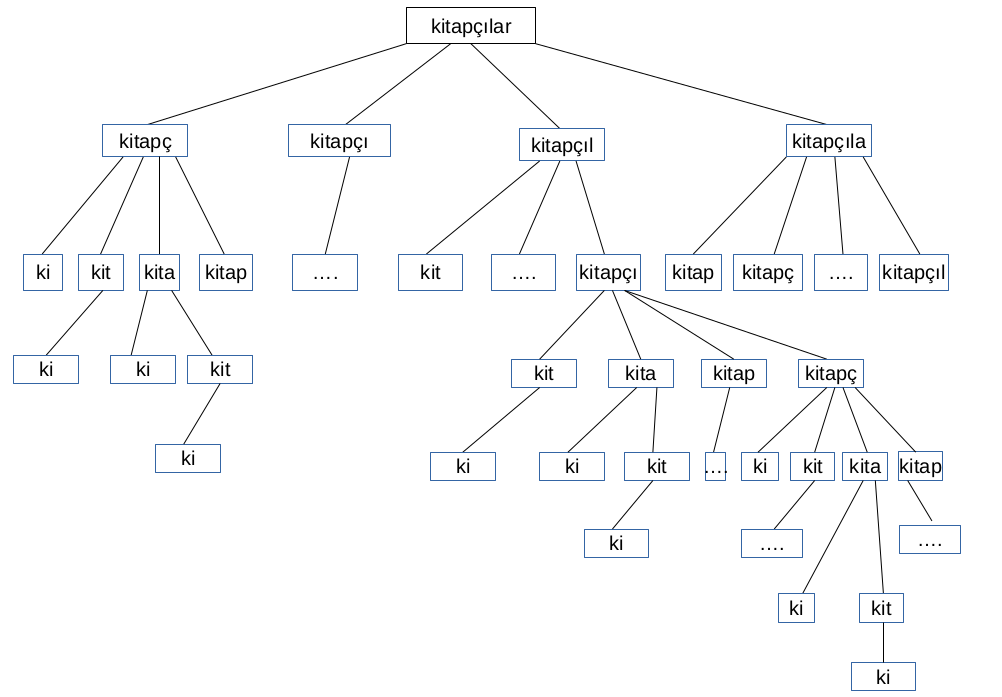}
  \caption{Recursive candidate generation of a Turkish word \textit{kitap\c{c}{\i}lar} (means \textit{bookshops}). For the simplicity, only the stems (left part of the binary segmentation) are shown in the segmentation tree. }
  \label{fig:wordtree}
\end{figure*}

An example is given in Figure~\ref{fig:wordtree}. The candidates of the word \textit{kitap\c{c}{\i}lar} (means \textit{bookshops}) is generated recursively. As can be seen in the figure, base word has at most 4 child nodes that corresponds to the first level candidates\footnote{We chose 4 because of the fact that the longest suffix in Turkish language is 4, e.g \textit{-iyor}.} (i.e. having a suffix with maximum 4 letters).

This recursive generation creates a large candidate space that also enlarges the observed space. The recursive procedure is given in Algorithm~\ref{algorithm}.

The model is learned by optimizing the feature weights according to the model likelihood given in Equation~\ref{likelihood}. Gradient-descent algorithm is used for the optimization similar to the original model. Once the model is learned, the prediction is performed for a novel word through the optimized weights for each feature, where again a recursive segmentation is applied.

\section{Results}
\label{results}

\begin{table}[t!]
\caption{Corpora size. MC: Morpho-challange. MC-05:10 Aggregated test data from Morpho Challange 2005--2010}
\begin{center}
    \begin{tabular}{ | l | l | l | l |} \hline
    \textbf{Language} & \textbf{Train} & \textbf{Test} &\textbf{WordVectors}\\ \hline
    Turkish &MC-2010 (617K) & MC-05:10 (2534) & Vectors-Gencor (361M)  \\ \hline
    English &MC-2010 (878K) & MC-05:10 (2218) & Wikipedia-Normalized (129M)  \\ \hline
    German  &MC-2010 (2M) & MC-2010 (785) & Manually collected (651M) \\ \hline
    \end{tabular}
    \label{datatable}
\end{center}
\end{table}

We use the publicly available datasets provided by Morpho Challenge~\cite{MorphoChallengeWeb} for both training and testing. The training sets contain 878K words, 617K words, and 2M for English, Turkish, and German respectively. The test sets contain 2200 words, 2500 words, and 785 words for English, Turkish, and German respectively that are also obtained from Morpho Challenge gold standard datasets by aggregating the gold sets in Morpho Challenge 2005-2010. 

We also use large datasets for training the neural word embedding model, word2vec~\cite{Mikolov} in order to build the neural word embeddings for the semantic similarity feature. All neural word embeddings are 200-dimensional. The corpora size is given in Table \ref{datatable}.



Experiments and evaluation are held as in the original paper. Segmentation points in the results are compared to those given in gold segmentation data, and Precision, Recall and F-1 measure values are calculated accordingly.

\begin{table}[t!]
\caption{Comparison of MorphoChains-R (with recursive candidate generation) with MorphoChains-O (the original MorphoChains system), Lee Segmenter, and other Morfessor members for Turkish, English and German}
\begin{center}
    \begin{tabular}{  l  l  l  l  p{1cm} } \hline
    \textbf{Language} & \textbf{Method} & \textbf{Precision} & \textbf{Recall} & \textbf{F-1}\\ \hline
    Turkish & MorphoChains-R  & 0.70 & 0.74 & \textbf{0.72} \\ 
    		& MorphoChains-O  & 0.49  & 0.76 & 0.60\\ 
            & Morfessor-CatMAP & 0.52 & 0.60 & 0.56 \\ 
            & Morfessor-Baseline & 0.82 & 0.36 & 0.50 \\  
            & Lee Segmenter &0.78  & 0.35 & 0.48  \\  \hline
    English & MorphoChains-R & 0.88 & 0.64 & \textbf{0.74} \\ 
    		& MorphoChains-O & 0.67 & 0.79 & 0.71 \\ 
            & Morfessor-Baseline & 0.74 & 0.62 & 0.67 \\  
            & Lee Segmenter & 0.82  & 0.52 & 0.64  \\  
            & Morfessor-CatMAP & 0.67 & 0.58 & 0.62 \\ \hline
    German  & Morfessor-Baseline & 0.55 & 0.54 & \textbf{0.54} \\
    		& MorphoChains-O & 0.33 & 0.49 & 0.38 \\
            & MorphoChains-R & 0.21 & 0.33 & 0.25 \\ \hline
    \end{tabular}
    \label{resulttable}
\end{center}
\end{table}

We compare our Turkish and English results with MorphoChains-O \cite{narasimhan2015unsupervised} (original MorphoChains system), Morfessor Baseline \cite{creutz2002unsupervised}, Morfessor CatMAP \cite{Creutz05inducingthe} and Lee Segmenter \cite{lee2011modeling}. All models are trained on the same train and test sets. Recursive candidate generation notably improves the scores with 12\% on Turkish with a final F-measure of 72\%, whereas the original MorphoChains system has a F-measure of 60\%. The same improvement also applies in English, having a F-measure of 74\% with 3\% improvement compared to the original MorphoChains system which has a F-measure of 71\%. 

We compare the recursive MorphoChains system with the original MorphoChains system and Morfessor Baseline on also German. Morfessor Baseline outperforms two other models with a F-measure of 54\%. The German results are better in the original model with a F-measure of 38\%, whereas the recursive model gives 25\%. This is possibly because of the morphological structure of the German language. German is not an agglutinative language and the left-recursion applied in the candidate generation will generate more erroneous candidate suffixes. This is also because of the common compounds in German language. All results for English, Turkish and German are given in Table~\ref{resulttable}. 
 
Results suggest that enlarging the candidate space will also enlarge the neighborhood size. Since contrastive estimation performs better on larger datasets, enlarging the size of the candidate space improves the precision scores because of the improved sub-word counts. For example, for the word \textit{kitap\c{c}{\i}lar}, the recursive candidate space will contain many valid candidates which in turn will help learning correct weights for correct candidates. 

\begin{table}[t!]
\caption{Example to correct and incorrect segmentations in the original (MorphoChains-O) and the recursive (MorphoChains-R) MorphoChains system}
\begin{center}
    \begin{tabular}{ l  l  l } \hline
    \textbf{Language} & \textbf{MorphoChains-O (Incorrect)} & \textbf{MorphoChains-R (Correct)} \\ \hline
    Turkish & s-\"{o}n-d-\"{u}-rme-ye & s\"{o}n-d\"{u}r-me-ye  \\ 
    		& cerrah-lara & cerrah-lar-a \\ 
            & b-a-\u{g}-lan-ma-mIz & ba\u{g}-lan-ma-mIz  \\ 
            & \"{o}\u{g}r-en-me-si-dir & \"{o}\u{g}ren-me-si-dir  \\ 
            & k{\i}-\c{s}-k-{\i}-r-t-ma-lar-In & k{\i}\c{s}k{\i}rt-ma-lar-{\i}n  \\ \hline
    English & sid-e-s-wipes & sides-wipe-s  \\ 
    		& mediterranea-n & mediterranean  \\ 
            & lef-t-'s & left-'s  \\
            & to-t-ed & tot-ed \\ 
            & pelle-t & pellet  \\ 
            & piz-za & pizza\\ \hline
    \end{tabular}
    \label{segmentationtable}
\end{center}
\end{table}

Some examples to correct and incorrect segmentations are given in Table \ref{segmentationtable}. In the original MorphoChains system, words are prone to be oversegmented, especially in Turkish.  In the recursive MorphoChains system, more words are segmented correctly by overcoming the oversegmentation problem in the original model. 

All this information can let us claim that increasing the candidate space in log-linear models improves the segmentation results especially in agglutinative languages such as Turkish. Enlarging the unobserved word space has been studied before via negative sampling. However, enlarging the obsverved space has not been studied before to our knowledge. In this paper, we show how it affects to enlarge the observed space in such a log-linear model. The results show that its affect is noticeably high. 

\section{Conclusion and Future Work}
\label{conclusion}
In this paper, we extend the unsupervised morphological segmentation system called MorphoChains \cite{narasimhan2015unsupervised}. We adopt the original log-linear model that uses contrastive estimation with negative sampling and aim to enlarge the observed space from which probability mass will be shifted.

We enlarge the observed candidate space by generating candidates recursively, whereas in the original model candidates are generated through binary segmentations of each word. Therefore, for each word the number of candidates is equal to the number of letters in each word. The recursion provides generating candidates that extract the suffixes in the middle of the word and this increases the probability assigned to these suffixes. However, in the original model only the probability of suffixes at the end of the words are increased with their occurrence counts in the corpus. 

We aim to try different optimization algorithms in the original log-linear model as a future goal. We believe that using a better optimization technique will also improve the results further. 

\section*{Acknowledgments}
This research is supported by the Scientific and Technological Research Council of Turkey (TUBITAK) with the project number EEEAG-115E464 and we are grateful to TUBITAK for their financial support.

%
%
\bibliographystyle{splncs}
\bibliography{lncs}

\end{document}